\documentclass[runningheads]{llncs}
\usepackage[T1]{fontenc}

\usepackage[backend=biber]{biblatex}
\usepackage{graphicx}

\usepackage{tabularx}
\usepackage{array}

\newcolumntype{Y}{>{\raggedright\arraybackslash}X}
\begin{document}
\raggedbottom
\title{AquiLLM: An Architecture for Supporting Tacit Knowledge Capture in Research Groups}
\titlerunning{AquiLLM Architecture for Tacit Knowledge Capture}

\author{
Jack Stark\inst{2}\orcidID{0009-0002-3049-3371} \and
Srinath Saikrishnan\inst{2}\orcidID{0009-0002-5638-0914} \and
Vikram Seenivasan\inst{2}\orcidID{0009-0006-5531-2126} \and
Bernie Boscoe\inst{1}\orcidID{0000-0001-6790-7297} \and
Andrew Lizarraga\inst{2} \and
Tuan Do\inst{2}
}
\authorrunning{J. Stark et al.}

\institute{
Southern Oregon University, Ashland, Oregon, USA\\
\email{boscoeb@sou.edu}
\and
University of California, Los Angeles, Los Angeles, California, USA\\
\email{jstark@astro.ucla.edu, srinathsai22@ucla.edu, vikrams25@g.ucla.edu, andrewlizarraga@ucla.edu, tdo@astro.ucla.edu}
}
\maketitle

\begin{abstract}
 Recent advances in retrieval-augmented generation (RAG) and large language models (LLMs) enable researchers to integrate AI into scientific workflows. However, using proprietary commercial AI systems raises concerns about transparency, reproducibility and privacy, which are essential for scientific practices. To this end, AquiLLM was developed as an open-source modular RAG-LLM framework using open-weight models, designed to support research groups in capturing tacit knowledge. In this work, we present a series of architectural improvements and feature enhancements to AquiLLM, including local embedding and reranking, multimodal capabilities, OpenAI-compatible inference interfaces, user interface improvements, semantic and episodic memory capabilities, and skills support. These enhancements were informed by discussions with domain experts, including astrophysicists and environmental researchers, and represent a step toward AI systems more closely aligned with scientific research practices.

\end{abstract}
\section{Introduction}
Recent advances in retrieval-augmented generation (RAG) and large language models (LLMs) have created new opportunities to integrate conversational AI systems into scientific research workflows. These tools can assist with tasks ranging from literature review and hypothesis formulation to code generation and data exploration. Yet, proprietary commercial services are often misaligned with scientific requirements of transparency, reproducibility, and data privacy.

At the same time, research is an inherently collaborative and iterative social practice. Scientific work depends not only on formal artifacts such as publications, but also on informal practices that emerge within research groups.

Much of this knowledge is tacit: it is developed through experience, embedded in workflows, and rarely fully documented \cite{collins2013, polanyi1967}. As a result, it is difficult to retain as group members come and go and project goals evolve. These challenges are particularly pronounced within small to midsize research groups, where knowledge sharing, infrastructure, and research workflows are often highly localized. However, few RAG-LLM systems are designed specifically for the workflow, privacy, and infrastructure requirements of such environments.
To this end, AquiLLM was developed as an open-source, modular RAG-LLM framework designed to support research groups in capturing and interacting with tacit knowledge \cite{campbell2025, aquillm2026, aquillmgithub2026}. As a customizable conversational AI system, AquiLLM can be tailored to research workflows and provides an interface comparable to commercial systems. Inference, retrieval, document processing, and memory are handled within locally controlled infrastructure, making the system suitable for environments in which researchers handle sensitive data. 

This work focuses on a series of technical enhancements to the base AquiLLM system motivated by scientific research needs and grounded in real-world deployment within an astrophysics research group at the University of California, Los Angeles. Through iterative engagement with domain scientists, including astrophysicists and environmental science researchers over two years, we identified limitations and opportunities related to usability, privacy, integration into research practices, and the capture and reuse of knowledge. The updates to AquiLLM include a revised architecture that includes local embedding and reranking, integration with efficient inference frameworks for open-weight models, and a configurable memory layer for persisting contextual information across sessions. Together, these enhancements support collaborative multimodal interaction over collections of documents, images, and other media while maintaining local control over research infrastructure and knowledge.

\section{Background}
The current state of academia-developed LLM platforms for research and education is best described as fragmented and evolving. Recent work has examined whether modern language models exhibit forms of reasoning or should instead be understood as purely statistical systems \cite{lizarraga2025}. At the same time, LLM-driven systems are inherently non-deterministic, complicating evaluation across scholarly criteria such as faithfulness, reproducibility, and epistemic reliability \cite{bender2021, ji2023}.

A number of institutional efforts illustrate both the promise and complexity of this space. Large-scale infrastructure initiatives, such as those at the Texas Advanced Computing Center (TACC), aim to integrate AI into research workflows \cite{tacc_nairr_2024}. Similar efforts by research institutions and organizations such as the Allen Institute for AI demonstrate growing investment in institutionally grounded AI ecosystems \cite{ai2_more_than_open}. Library-led efforts, including Carnegie Mellon University's Open Forum for AI (OFAI), further highlight the role of scholarly infrastructure in evaluating and mediating AI tools for research use \cite{cmu_ofai}.

At the same time, there is increasing demand for locally deployable or institutionally controlled systems that do not depend on commercial providers. Conversations with researchers and developers consistently highlight concerns that submitted data may be retained or used for training \cite{bommasani2022, yuan2025, phan2026}. Transparency is also a growing issue, including questions about training data, model behavior, and output generation. 

In response, many research-related platforms are designed to allow adopters to swap commercial models for open-weight or locally hosted alternatives, provided sufficient computational resources are available \cite{yang2025}.

Over time, as AquiLLM was introduced and adopted across research groups, additional challenges emerged. These included the need to transition toward open-weight models and locally hosted infrastructure, as well as growing user demand for multimodal capabilities. The move toward local, open-weight deployment reflected a broader need for greater control over how the system handled research data, integrated with existing infrastructure, and could be maintained over time. In particular, researchers expressed interest in tools that could read plots, interpret figures, and engage with visual data in the context of meeting slides or publications. There was also an increasing need for more flexible system integration, enabling AquiLLM to adapt to diverse workflows, data types, and disciplinary practices while maintaining alignment with research norms. These requirements also shaped how the system should be evaluated.

Since AquiLLM is intended for collaborative scientific settings rather than general-purpose consumer use, conventional LLM benchmarks only partially capture the challenges relevant to deployment. It has been shown that models that perform strongly on standard retrieval benchmarks may still struggle to faithfully use localized scientific knowledge, particularly in contexts requiring ambiguity resolution, domain-specific reasoning, or adherence to retrieved research materials \cite{boscoe2026aquillm}. As a result, we assess the proposed system enhancements through domain-specific research use cases rather than benchmark performance alone.

Evidence for these claims comes from a prior domain-expert evaluation of AquiLLM conducted within an astronomy research group. In that study, five astronomers affiliated with the same research group evaluated the faithfulness of 141 responses generated by AquiLLM over a curated corpus of 31 astronomy documents under a no-internet-use constraint \cite{boscoe2026aquillm}. The evaluation included questions requiring factual retrieval, scientific interpretation, comparative analysis, metadata identification, and conceptual or onboarding knowledge, allowing the system to be examined across different forms of scientific information seeking. Experts assessed response quality and accuracy relative to the source literature available to the system.

Responses to questions involving scientific interpretation and factual retrieval received the highest ratings, indicating that the system performed well when relevant evidence could be identified and grounded within a relatively localized portion of the corpus. Comparative scientific analysis produced the lowest average rating, particularly when answering a question required information to be retrieved and synthesized across multiple documents. Expert review identified several recurring failure modes, including unsupported synthesis across fragmented sources, retrieval failures, subtle factual errors, overconfident analysis, and inconsistencies across repeated evaluations. Importantly, these failures were not limited to the underlying language model: several arose from interactions among retrieval, context selection, evidence availability, and response generation.
These findings suggest that locally controlled, open-weight RAG architectures can support interpretable and faithful use of specialized scientific documents, particularly when supporting evidence is clearly represented and strongly localized within the available collection. At the same time, the evaluation exposes limitations that become increasingly important as research collections grow and questions require information to be integrated across sources or interactions. These results motivate architectural mechanisms for improving retrieval-grounded response generation, maintaining efficient and inspectable provenance, automatically constructing relevant context from distributed evidence, and calibrating the reuse of prior conversational and research context. In AquiLLM, these mechanisms are intended not only to improve individual responses, but also to support the longer-term accumulation and reuse of knowledge within collaborative scientific work.

\begin{table}[t]
\centering
\small

\caption{Comparison of collaborative knowledge organization across AquiLLM, Open WebUI, and NotebookLM.}
\label{tab:collaborative-knowledge-comparison}

\begin{tabularx}{\linewidth}{@{}YYYY@{}}
\hline
\textbf{Dimension} & \textbf{Open WebUI} & \textbf{NotebookLM} & \textbf{AquiLLM} \\
\hline

Collaboration model
& Shared resources
& Shared notebooks
& Shared research workspace \\

Shared knowledge
& Knowledge bases
& Notebook sources
& Project collections \\

Group context
& Resource-level
& Notebook-level
& Project/group-level \\

Persistent context
& No unified group context
& Notebook context
& User memory + project collections \\

Tacit/procedural knowledge
& General knowledge resources
& Source-based
& Explicit focus \\

Deployment
& Self-hosted
& Cloud-managed
& Self-hosted \\

\hline
\end{tabularx}
\end{table}
Although tools such as ChatGPT, Claude, and Google NotebookLM are widely used for research support, reliance on commercial APIs and externally managed services introduces practical concerns for some research environments. As Yang notes, these systems can incur significant financial costs and raise issues related to data privacy, security, changing API interfaces, and dependence on external infrastructure \cite{yang2025}. These constraints are particularly relevant for researchers working with sensitive or restricted data, limited computational budgets, or requirements for greater control over how research materials are stored and processed. AquiLLM addresses these concerns through a full-stack architecture designed for locally hosted, researcher-controlled infrastructure, allowing research groups to retain control over their data, models, retrieval pipeline, and computational resources.

Local control alone, however, does not address how knowledge is organized and maintained within a research collaboration. Table~\ref{tab:collaborative-knowledge-comparison} compares AquiLLM with Open WebUI and NotebookLM, two systems that also support source-grounded LLM interactions and forms of collaborative access, but structure that collaboration differently. Open WebUI primarily supports collaboration through access to shared resources, such as knowledge bases and tools, while NotebookLM enables users to share individual notebooks and their associated sources. AquiLLM instead organizes collaboration around a persistent research group or project, providing a shared context in which collections, evidence, procedures, and accumulated group knowledge can be maintained and reused over time. This design reflects AquiLLM's emphasis on knowledge continuity within long-lived scientific collaborations rather than document sharing or individual interactions alone.

\section{Improvements to AquiLLM}
We designed AquiLLM improvements around a core constraint common to research environments: systems must handle heterogeneous document workflows while keeping cost and latency predictable under limited compute resources. These goals motivate design decisions emphasizing modularity, local deployment, and explicit control of data, retrieval, and persistent system state.

We deploy AquiLLM on Jetstream2 using separate development and production environments to reflect real-world usage \cite{hancock2021}. Both environments run on g5.xl instances equipped with NVIDIA H100 GPUs for local model inference, with attached 500 GB volumes for persistent storage of documents, indexes, and intermediate artifacts. Each instance provides 20 vCPUs and 240 GB of system RAM, supporting multiple users and services. A key constraint is GPU memory (VRAM), which must hold both model weights and the key-value (KV) cache used during inference. The KV cache grows with the sequence length and concurrent requests. When VRAM is insufficient, context length or concurrency must be reduced, or portions of the cache must be offloaded to system memory, introducing additional latency. System RAM provides headroom for these fallback strategies, but performance remains bounded by GPU memory capacity.

\begin{figure}[t]
  \centering
  \includegraphics[width=\linewidth]{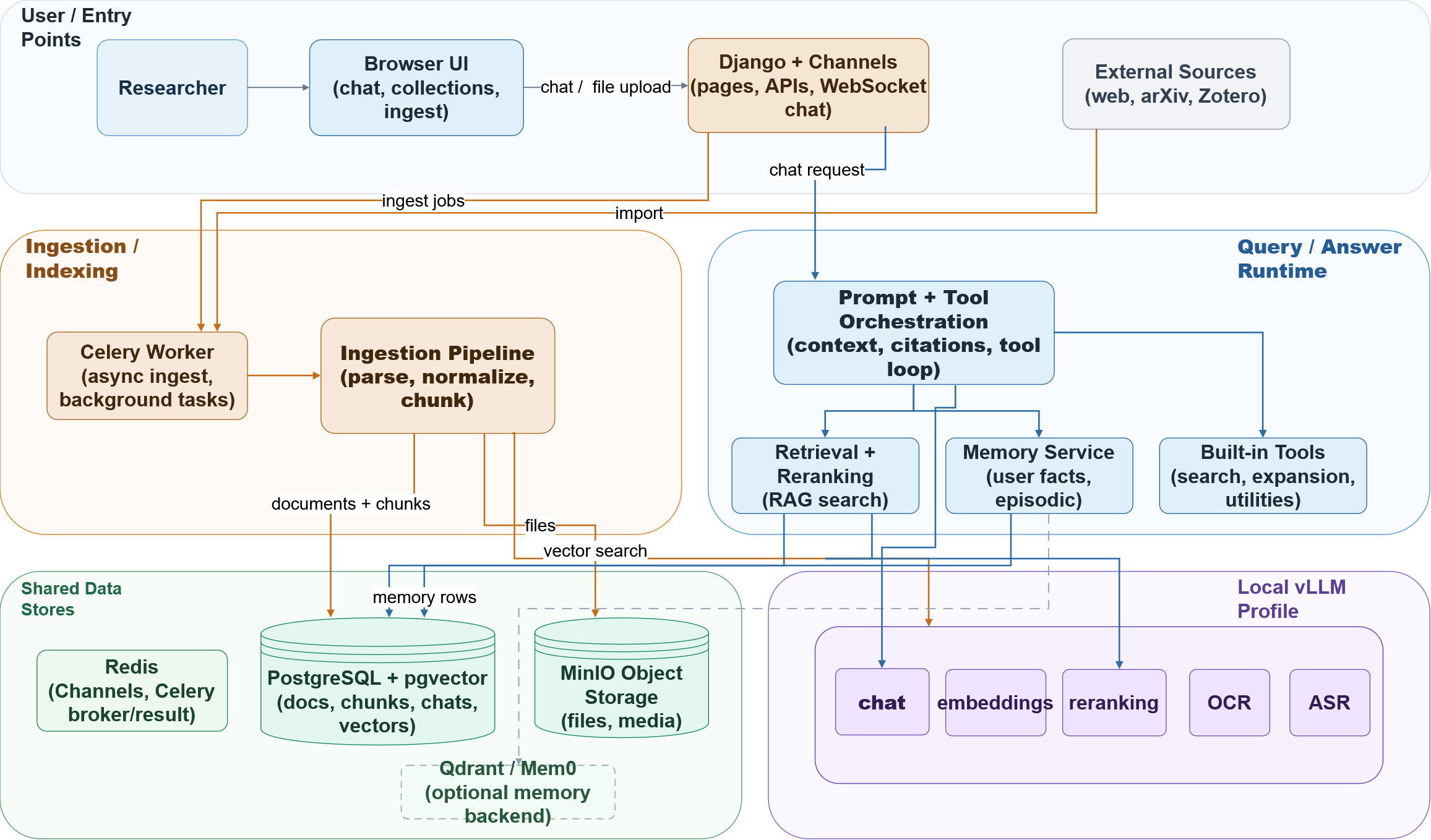}
  \caption{Architecture of AquiLLM. Background workers handle document parsing and indexing into shared storage, while runtime requests pass through a prompt orchestration layer coordinating retrieval, memory, and tool use. Model inference is served locally via vLLM with separate components for chat, embedding, reranking, and multimodal processing.}
  \label{fig:aquillm-arch}
\end{figure}

One of the first major decisions was to modularize the backend into domain-scoped applications rather than maintaining a monolithic codebase (see Fig.~\ref{fig:aquillm-arch}). The figure shows the separation between the ingestion, retrieval, orchestration, and inference components. Ingestion and indexing are handled asynchronously through background workers, while query-time execution flows through a prompt orchestration layer that coordinates retrieval, memory, and tool use. Model inference is decoupled into a local serving layer, enabling different models for chat, embedding, reranking, and multimodal tasks to operate independently. In practice, the original structure made it difficult to isolate changes, as ingestion, retrieval, and model interaction were tightly coupled. Separating these into modules reduced unintended interactions and allowed independent iteration on system components such as reranking and memory. This separation also made resource use tunable at each boundary: retrieval can reuse cached embeddings and rerank results; the orchestration layer applies explicit token budgets to tool evidence, episodic memory, and conversation history before inference; and the serving layer can host smaller specialized models for embedding and reranking rather than routing all workloads through the chat model. These controls are optional and configuration-dependent, but they bound worst-case prompt growth and redundant retrieval work on constrained GPUs; per-request token and latency accounting is logged for rollout validation rather than reported here as a single end-to-end benchmark. 

We treat local model execution as a pluggable inference capability and standardize calls through the OpenAI-compatible HTTP interface provided by vLLM. vLLM is an inference engine for serving open-weight models that handles batching, scheduling, and memory management, allowing multiple requests to share GPU resources efficiently \cite{kwon2023}. This comes with operational cost: batching improves throughput but can increase latency variability under mixed workloads, and each deployment must tune GPU memory, concurrency, and context limits. Multimodal support further increases this pressure because embedding, reranking, transcription, and chat inference may all compete for compute at different points in the request path. AquiLLM therefore uses the vLLM interface not only as a serving layer, but also as an abstraction boundary between the application and specialized inference components.

This abstraction provides a consistent invocation model for both interactive requests and background processing while allowing workload specialization through sidecar models for chat, embedding, reranking, transcription, and multimodal processing. We mitigate the resulting resource pressure at both the serving and application layers. At the serving layer, we cap per-service GPU memory and sequence concurrency, isolate sidecar workloads, and start services serially to reduce startup contention; optional LMCache wiring supports KV offloading when enabled. At the application layer, short-lived exact-match caches store repeated retrieval artifacts such as query embeddings, rerank outputs, document lookups, permission-filtered access checks, and image payloads. These caches reduce redundant inference and database work on follow-up turns and repeated searches within a session. Token-budget controls bound prompt growth, and direct document questions are routed through a direct RAG path to avoid unnecessary chat turns. These mechanisms are designed to fail open, so inference degrades gracefully rather than failing hard when a sidecar model or cache path is unavailable.

On our production-like development deployment, we use Qwen3.6-27B as our default chat model \cite{qwen_qwen36_27b_2026}. This model was chosen for its strong reasoning, coding, and tool-use capabilities for a model of its size, and as a hybrid gated delta network it is well suited to long-context workflows. Its multimodal capabilities also support day-to-day research use within AquiLLM. In AquiLLM, both the chat-template and inference pipelines can pass uploaded documents through the model’s multimodal capabilities during chat interactions. This allows the default chat model to use text and image data from user-uploaded materials while also supporting the RAG pipeline without requiring a separate OCR container or sidecar. Whisper-large-v3-turbo \cite{radford_robust_2022} is used for transcription because it provides a strong balance between ASR performance and compute overhead.

We also include Qwen3-VL-Embedding-2B and Qwen3-VL-Reranker-2B to support multimodal embedding and reranking within the retrieval pipeline, following broader work on sentence embedding and reranking for retrieval \cite{pandit2025, reimers2019}. These components are intended to improve retrieval quality without increasing prompt size, though they have not yet been fully evaluated in this deployment. The same chat and embedding models are used within our vector database-backed memory system implementation to maintain consistency in tokenization and embedding geometry. Finally, we optionally deploy an LLMLingua-2 prompt-compression checkpoint, llmlingua-2-xlm-roberta-large-meetingbank,
on CPU, following prior work on prompt compression for accelerated inference \cite{jiang_llmlingua_2023}. This trades additional CPU overhead for reduced token usage and increased effective context length. On the retrieval side, replacing external services such as Cohere with local embedding and reranking improves privacy and provides greater control over the retrieval pipeline \cite{gao2024}.

\begin{table}[t]
\centering
\small
\setlength{\tabcolsep}{4pt}
\renewcommand{\arraystretch}{1.15}

\caption{Inference capabilities in a representative local deployment.}
\label{tab:inference-capabilities}

\begin{tabular}{p{0.20\linewidth} p{0.27\linewidth} p{0.43\linewidth}}
\hline
\textbf{Capability} & \textbf{Engine} & \textbf{Notes} \\
\hline
Chat & vLLM or hosted API & Selected at deploy time \\
OCR & Qwen multimodal OCR or legacy sidecar & Tesseract fallback available \\
Transcription & Whisper via vLLM & Audio and video ingestion \\
Embedding & vLLM pooling runner & Cohere fallback for text only \\
Reranking & vLLM HTTP rerank & Improves retrieval precision \\
Prompt compression & LLMLingua-2 on CPU & Optional; disabled by default \\
\hline
\end{tabular}
\end{table}

To augment retrieval, we add a configurable memory component to AquiLLM, allowing representation of user-specific, long-horizon state not captured by documents in the corpus. This is maintained outside of any collection and can be mutated over time \cite{maharana_evaluating_2024}. We treat two broad categories of memory: stable facts about the user encode preferences, tone, goals, and projects, providing static context, while episodic semantic memories contain information retrieved from previous conversations and are indexed to allow similarity search against prior message history. By default, reads and writes to memory are backed by PostgreSQL tables with pgvector extensions. With Mem0 connected, episodic memories can be forwarded to the Mem0 SDK to be stored in Qdrant along with configurable graph backends, falling back to dual-write into local tables. Because memory can accumulate long-lived, user-specific state, it is treated as private, deployment-controlled data rather than as part of a shared document corpus. Memory is disabled by default and must be explicitly configured for a deployment. When enabled, memory uses the same chat and embedding models within Mem0, allowing us to maintain the same tokenization and embedding geometry while preserving local control over memory storage and retrieval. This also introduces additional complexity around privacy and correctness, such as stale memories, so the functionality remains optional and configurable. At retrieval time, if possible, the current conversation is omitted to reduce immediate self-echo, and external memory errors fail open to local retrieval.

To work alongside the semantic memory layer, we have introduced skills and skill packs which are exposed to end users through collections. These operate as a form of procedural memory \cite{sumers2024coala}. Through these skills and skill packs, AquiLLM allows research teams to associate directed workflow instructions with their corpus collections, which already define the boundaries of their documents. When someone opens a chat with a project collection, Markdown prompts such as onboarding documents, plotting standards, reduction checklists, or subfield-specific analysis workflows are fetched and loaded into the session prompt of the chat interface. This happens seamlessly without modification of fundamental platform code or making these instructions visible to anyone working on other projects. 

A single document may have one narrow ability. A sub-collection may define an entire skill set for a lab or instrument pipeline. Because prompts load after the collection is chosen and access permissions are verified, implicit knowledge is bundled with the knowledge it describes: a spectroscopy team and a photometry team can each have their own corpus with entirely separate "how to work in this group" instructions, maintaining separate retrieval behavior and default actions. These collection skills direct the manner in which the model reads and uses retrieved information during that conversation; they are not additional functions. Skills consume context when their associated collection is active, but remain scoped to that collection and are not introduced into conversations associated with other projects or research groups. Applied to science, this allows teams to bridge the gap between the contents of what is in a collection and the norms with which a team would like that content to be interpreted, referenced, and leveraged.

To help make better use of our other improvements, we introduce salience-aware context packing into the generation pipeline. Instead of heuristic truncation, we utilize deterministic token budgets with pinned sections and staged pruning of imported context with overflow fallbacks to provide better token efficiency and determinism when faced with limited GPU resources \cite{jiang_llmlingua_2023}. The available prompt window is partitioned into history and tool-evidence budgets; unpinned messages are ranked by lexical overlap with the latest user query, citation/entity boosts, and recency, then pruned in a fixed order: tool-header deduplication, boilerplate collapse, extractive shortening, and selective LLMLingua-2 compression of below-median-salience content, followed by salience-ordered dropping and marked hard trimming. Risks of salience heuristics missing edge-case content are mitigated by guaranteeing inclusion of known critical content---the latest turn, contiguous tool-call/result chains, and the last \(N\) primary user turns---while preserving overall provenance via retained message roles and explicit truncation markers.

Finally, we pair these backend changes with usability improvements. The ingestion process has been overhauled to take advantage of concurrent processing on a per-file basis (task fan-out), reducing time to first result for larger collections of data. There are natural burstiness costs to this approach, so we optimized the surrounding batch system to decrease per-use cost under load \cite{kwon2023}. Lastly, we iterated on the UI to improve overall workflow and readability. We added KaTeX rendering support to give the model the ability to properly display equations in its output. We complement this with a newly polished chat experience to make the end user experience more pleasant for scientific and serious workflows.

\section{Limitations}
A number of notable shortcomings remain in AquiLLM as it stands today. First, the codebase retains a hybrid architecture while older compatibility modules are incrementally migrated into domain-first apps. This creates provider-behavior asymmetries between hosted and local backends, especially around streaming, message trimming, and request options. Second, writes to episodic memory are asynchronous and eventually consistent. This keeps the UI responsive, but can create a short-lived mismatch between a recent interaction and what can be retrieved from memory. Third, many features remain behind flags and require deployment-specific initialization and verification beyond the baseline profile. This includes hosted API access for chat models and fallback embedders, which are available by default but can be disabled for fully local deployments. Fully air-gapped deployments will need to be explicitly configured to keep inference, embedding, and retrieval all on-device. Fourth, our UX presently consists of Django-rendered pages with React islands. While this allows us to incrementally build the interface, it does not fully match the interaction patterns set by popular consumer chat products. Finally, retrieval speed and document-scaling characteristics vary based on corpus size, available compute, and indexing strategy. We have evaluated AquiLLM's faithfulness on formal astronomy knowledge bases; however, we have not yet benchmarked our current approach on informal knowledge artifacts produced collaboratively by research groups such as meeting minutes, Slack messages, code comments, or collection-specific playbooks. Therefore, rigorously evaluating the capture and reuse of tacit knowledge remains to be done \cite{boscoe2026aquillm}.

\section{Future Work}
Several directions emerge from this work. A near-term priority is improving prompt compression. Our current deployment uses LLMLingua-2 on CPU to reduce token usage, but its MLP-based design introduces latency. Recent MLP-free approaches offer a path toward lower-latency compression better suited to real-time inference \cite{honig2025}.
A second direction is expanding support for domain-specific scientific modalities. Research groups in astrophysics routinely work with data types such as spectra and photometric distributions that are not well represented in standard LLM interfaces. Multimodal masked autoencoder architectures trained on image-spectrum associations \cite{himes2025} suggest a path toward embeddings that better preserve scientific structure. Incorporating these into the retrieval pipeline could improve relevance for domain-specific workflows.
Looking ahead, the current memory captures user-level state but does not support structured planning over longer horizons. Latent planning approaches \cite{noh2025, kong2025} suggest that learned representations can support goal-directed behavior. Extending these ideas to research workflows could enable AquiLLM to move beyond reactive retrieval toward more proactive knowledge support.

\section{Conclusion}
We presented an updated version of AquiLLM, a modular RAG-LLM system designed for research groups. The system emphasizes local deployment with open-weight models and supports both formal and informal knowledge within collaborative workflows. Through local embedding and reranking, efficient inference, a memory layer, and salience-aware context construction, AquiLLM supports document-grounded interaction under practical constraints.
A central focus is the capture and reuse of tacit knowledge. By supporting multimodal ingestion of informal materials, AquiLLM better reflects how research is conducted. These design choices introduce additional complexity but are necessary to support reproducibility within research infrastructure. Extending AquiLLM to support longer-horizon reasoning remains an open challenge and a direction for future work.

\printbibliography

@inproceedings{hancock2021,
	address = {Boston MA USA},
	title = {Jetstream2: {Accelerating} cloud computing via {Jetstream}},
	isbn = {978-1-4503-8292-2},
	shorttitle = {Jetstream2},
	url = {https://dl.acm.org/doi/10.1145/3437359.3465565},
	doi = {10.1145/3437359.3465565},
	language = {en},
	urldate = {2024-07-09},
	booktitle = {Practice and {Experience} in {Advanced} {Research} {Computing}},
	publisher = {ACM},
	author = {Hancock, David Y. and others},
	month = jul,
	year = {2021},
	pages = {1--8},
}

@book{polanyi1967,
	series = {Anchor books},
	title = {The {Tacit} {Dimension}},
	isbn = {978-0-14-341418-6},
	url = {https://books.google.com/books?id=jwLXAAAAMAAJ},
	publisher = {Anchor Books},
	author = {Polanyi, M.},
	year = {1967},
}

@book{collins2013,
	address = {Chicago, Ill},
	title = {Tacit and explicit knowledge},
	isbn = {978-0-226-00421-1 978-0-226-11382-1},
	language = {eng},
	publisher = {University of Chicago Press},
	author = {Collins, Harry},
	year = {2013},
}

@inproceedings{campbell2025,
	
	title = {{AquiLLM}: a {RAG} {Tool} for {Capturing} {Tacit} {Knowledge} in {Research} {Groups}},
	shorttitle = {{AquiLLM}},
	url = {http://arxiv.org/abs/2508.05648},
	doi = {10.48550/arXiv.2508.05648},
	urldate = {2025-10-25},
	publisher = {US-RSE Conference},
	author = {Campbell, Chandler and Boscoe, Bernie and Do, Tuan},
	month = oct,
	year = {2025},
	note = {arXiv:2508.05648 [cs]},
	}

@misc{gao2024,
	title = {Retrieval-{Augmented} {Generation} for {Large} {Language} {Models}: {A} {Survey}},
	shorttitle = {Retrieval-{Augmented} {Generation} for {Large} {Language} {Models}},
	url = {http://arxiv.org/abs/2312.10997},
	doi = {10.48550/arXiv.2312.10997},
	urldate = {2026-03-24},
	publisher = {arXiv},
	author = {Gao, Yunfan and others},
	month = mar,
	year = {2024},
	note = {arXiv:2312.10997 [cs]},
	}

@misc{kwon2023,
	title = {Efficient {Memory} {Management} for {Large} {Language} {Model} {Serving} with {PagedAttention}},
	url = {http://arxiv.org/abs/2309.06180},
	doi = {10.48550/arXiv.2309.06180},
	urldate = {2026-03-24},
	publisher = {arXiv},
	author = {Kwon, Woosuk and others},
	month = sep,
	year = {2023},
	note = {arXiv:2309.06180 [cs]},
	}

@misc{reimers2019,
	title = {Sentence-{BERT}: {Sentence} {Embeddings} using {Siamese} {BERT}-{Networks}},
	shorttitle = {Sentence-{BERT}},
	url = {http://arxiv.org/abs/1908.10084},
	doi = {10.48550/arXiv.1908.10084},
	publisher = {arXiv},
	author = {Reimers, Nils and Gurevych, Iryna},
	month = aug,
	year = {2019},
	note = {arXiv:1908.10084 [cs]},
	}

@misc{pandit2025,
	title = {The {Evolution} of {Reranking} {Models} in {Information} {Retrieval}: {From} {Heuristic} {Methods} to {Large} {Language} {Models}},
	shorttitle = {The {Evolution} of {Reranking} {Models} in {Information} {Retrieval}},
	url = {http://arxiv.org/abs/2512.16236},
	doi = {10.48550/arXiv.2512.16236},
	publisher = {arXiv},
	author = {Pandit, Tejul and Mahendru, Sakshi and Raval, Meet and Upadhyay, Dhvani},
	month = dec,
	year = {2025},
	note = {arXiv:2512.16236 [cs]},
}

@misc{yang2025,
	title = {Large language models in materials science and the need for open-source approaches},
	url = {http://arxiv.org/abs/2511.10673},
	doi = {10.48550/arXiv.2511.10673},
	publisher = {arXiv},
	author = {Yang, Fengxu and Chen, Weitong and Evans, Jack D.},
	month = nov,
	year = {2025},
	note = {arXiv:2511.10673 [cs]},
	}

@article{lizarraga2025,
  author  = {Lizarraga, Andrew and Honig, Edouardo and Wu, Ying Nian},
  title   = {From Stochastic Parrots to Digital Intelligence: The Evolution of Language Models and Their Cognitive Capabilities},
  journal = {WIREs Computational Statistics},
  volume  = {17},
  number  = {3},
  pages   = {e70035},
  year    = {2025},
  doi     = {10.1002/wics.70035},
  url     = {https://onlinelibrary.wiley.com/doi/abs/10.1002/wics.70035}
}

@inproceedings{bender2021,
	address = {New York, NY, USA},
	series = {{FAccT} '21},
	title = {On the {Dangers} of {Stochastic} {Parrots}: {Can} {Language} {Models} {Be} {Too} {Big}? },
	isbn = {978-1-4503-8309-7},
	shorttitle = {On the {Dangers} of {Stochastic} {Parrots}},
	url = {https://dl.acm.org/doi/10.1145/3442188.3445922},
	doi = {10.1145/3442188.3445922},
	booktitle = {Proceedings of the 2021 {ACM} {Conference} on {Fairness}, {Accountability}, and {Transparency}},
	publisher = {Association for Computing Machinery},
	author = {Bender, Emily M. and Gebru, Timnit and McMillan-Major, Angelina and Shmitchell, Shmargaret},
	month = mar,
	year = {2021},
	pages = {610--623},
}

@article{ji2023,
	title = {Survey of {Hallucination} in {Natural} {Language} {Generation}},
	volume = {55},
	issn = {0360-0300},
	url = {https://dl.acm.org/doi/10.1145/3571730},
	doi = {10.1145/3571730},
	number = {12},
	urldate = {2026-03-26},
	journal = {ACM Comput. Surv.},
	author = {Ji, Ziwei and others},
	month = mar,
	year = {2023},
	pages = {248:1--248:38},
}

@misc{bommasani2022,
	title = {On the {Opportunities} and {Risks} of {Foundation} {Models}},
	url = {http://arxiv.org/abs/2108.07258},
	doi = {10.48550/arXiv.2108.07258},
	publisher = {arXiv},
	author = {Bommasani, Rishi and others},
	month = jul,
	year = {2022},
	note = {arXiv:2108.07258 [cs]},

}

@misc{yuan2025,
	title = {The {Science} of {Evaluating} {Foundation} {Models}},
	url = {http://arxiv.org/abs/2502.09670},
	doi = {10.48550/arXiv.2502.09670},
	publisher = {arXiv},
	author = {Yuan, Jiayi and Zhang, Jiamu and Wen, Andrew and Hu, Xia},
	month = feb,
	year = {2025},
	note = {arXiv:2502.09670 [cs]},
}

@article{phan2026,
	title = {A benchmark of expert-level academic questions to assess {AI} capabilities},
	volume = {649},
	copyright = {2026 The Author(s)},
	issn = {1476-4687},
	url = {https://www.nature.com/articles/s41586-025-09962-4},
	doi = {10.1038/s41586-025-09962-4},
	language = {en},
	number = {8099},
	urldate = {2026-03-26},
	journal = {Nature},
	publisher = {Nature Publishing Group},
	author = {Phan, Long and others},
	month = jan,
	year = {2026},
	pages = {1139--1146},
}

@misc{honig2025,
	title = {Better {Prompt} {Compression} {Without} {Multi}-{Layer} {Perceptrons}},
	url = {http://arxiv.org/abs/2501.06730},
	doi = {10.48550/arXiv.2501.06730},
	publisher = {arXiv},
	author = {Honig, Edouardo and Lizarraga, Andrew and Zhang, Zijun Frank and Wu, Ying Nian},
	month = jan,
	year = {2025},
	note = {arXiv:2501.06730 [cs]},
	}

@misc{himes2025,
	title = {Multi-{Modal} {Masked} {Autoencoders} for {Learning} {Image}-{Spectrum} {Associations} for {Galaxy} {Evolution} and {Cosmology}},
	url = {http://arxiv.org/abs/2510.22527},
	doi = {10.48550/arXiv.2510.22527},
	publisher = {arXiv},
	author = {Himes, Morgan and Krishnamurthy, Samiksha and Lizarraga, Andrew and Saikrishnan, Srinath and Seenivasan, Vikram and Soriano, Jonathan and Wu, Ying Nian and Do, Tuan},
	month = oct,
	year = {2025},
	note = {arXiv:2510.22527 [astro-ph]},
	}

@inproceedings{noh2025,
	title = {Latent {Adaptive} {Planner} for {Dynamic} {Manipulation}},
	issn = {2640-3498},
	url = {https://proceedings.mlr.press/v305/noh25a.html},
	language = {en},
	urldate = {2026-03-28},
	booktitle = {Proceedings of {The} 9th {Conference} on {Robot} {Learning}},
	publisher = {PMLR},
	author = {Noh, Donghun and Kong, Deqian and Zhao, Minglu and Lizarraga, Andrew and Xie, Jianwen and Wu, Ying Nian and Hong, Dennis},
	month = oct,
	year = {2025},
	pages = {2430--2448},
	}

@misc{kong2025,
	title = {Latent {Plan} {Transformer} for {Trajectory} {Abstraction}: {Planning} as {Latent} {Space} {Inference}},
	shorttitle = {Latent {Plan} {Transformer} for {Trajectory} {Abstraction}},
	url = {http://arxiv.org/abs/2402.04647},
	doi = {10.48550/arXiv.2402.04647},
	publisher = {arXiv},
	author = {Kong, Deqian and Xu, Dehong and Zhao, Minglu and Pang, Bo and Xie, Jianwen and Lizarraga, Andrew and Huang, Yuhao and Xie, Sirui and Wu, Ying Nian},
	month = aug,
	year = {2025},
	note = {arXiv:2402.04647 [cs]},
}

@misc{qwen_qwen36_27b_2026,
  title        = {{Qwen3.6-27B}},
  author       = {{Qwen Team}},
  year         = {2026},
  month        = apr,
  publisher    = {Hugging Face},
  url          = {https://huggingface.co/Qwen/Qwen3.6-27B},
  note         = {Model weights and configuration files for the post-trained Qwen3.6-27B model}
}

@misc{radford_robust_2022,
	title = {Robust {Speech} {Recognition} via {Large}-{Scale} {Weak} {Supervision}},
	url = {https://arxiv.org/abs/2212.04356v1},
	
	language = {en},
	urldate = {2026-03-29},
	journal = {arXiv.org},
	author = {Radford, Alec and Kim, Jong Wook and Xu, Tao and Brockman, Greg and McLeavey, Christine and Sutskever, Ilya},
	month = dec,
	year = {2022},
	}

@misc{maharana_evaluating_2024,
	title = {Evaluating {Very} {Long}-{Term} {Conversational} {Memory} of {LLM} {Agents}},
	url = {http://arxiv.org/abs/2402.17753},
	doi = {10.48550/arXiv.2402.17753},
	publisher = {arXiv},
	author = {Maharana, Adyasha and Lee, Dong-Ho and Tulyakov, Sergey and Bansal, Mohit and Barbieri, Francesco and Fang, Yuwei},
	month = feb,
	year = {2024},
	note = {arXiv:2402.17753 [cs]},
	
}

@misc{jiang_llmlingua_2023,
	title = {{LLMLingua}: {Compressing} {Prompts} for {Accelerated} {Inference} of {Large} {Language} {Models}},
	shorttitle = {{LLMLingua}},
	url = {https://arxiv.org/abs/2310.05736v2},
	language = {en},
	journal = {arXiv.org},
	author = {Jiang, Huiqiang and Wu, Qianhui and Lin, Chin-Yew and Yang, Yuqing and Qiu, Lili},
	month = oct,
	year = {2023},
}

@inproceedings{boscoe2026aquillm,
author = {Boscoe, Bernie and others},
title = {{AquiLLM}: Evaluating Faithfulness in Open-Weight {RAG-LLM} Systems for Scientific Research},
booktitle = {Proceedings of the 2026 US Research Software Engineer Conference (US-RSE)},
year = {2026},
address = {San Jose, California},
note = {Accepted}
}

@article{sumers2024coala,
  title   = {Cognitive Architectures for Language Agents},
  author  = {Sumers, Theodore R. and Yao, Shunyu and Narasimhan, Karthik and Griffiths, Thomas L.},
  journal = {Transactions on Machine Learning Research},
  year    = {2024},
  url     = {https://openreview.net/forum?id=1i6ZCvflQJ},
  note    = {Also available as arXiv:2309.02427}
}

@online{tacc_nairr_2024,
  author  = {{Texas Advanced Computing Center}},
  title   = {{NSF Selects TACC Supercomputers for National AI Research Resource Pilot}},
  year    = {2024},
  month   = feb,
  day     = {1},
  url     = {https://tacc.utexas.edu/news/latest-news/2024/02/01/nsf-selects-tacc-supercomputers-for-national-ai-research-resource-nairr-pilot/},
  urldate = {2026-06-20}
}

@online{ai2_more_than_open,
  author  = {{Allen Institute for AI}},
  title   = {{More Than Open}},
  url     = {https://allenai.org/more-than-open},
  urldate = {2026-06-20}
}

@online{cmu_ofai,
  author  = {{Carnegie Mellon University}},
  title   = {{Open Forum for AI (OFAI)}},
  url     = {https://www.cmu.edu/engin/programs/ofai.html},
  urldate = {2026-06-20}
}

@misc{aquillm2026,
  title        = {AquiLLM},
  author       = {{AquiLLM}},
  year         = {2026},
  howpublished = {\url{https://aquillm.org/}},
  note         = {Accessed: 2026-06-21}
}

@misc{aquillmgithub2026,
  title        = {AquiLLM},
  author       = {{AquiLLM}},
  year         = {2026},
  howpublished = {\url{https://github.com/AquiLLM/AquiLLM}},
  note         = {GitHub repository. Accessed: 2026-06-21}
}

\begin{credits}

\subsubsection{\ackname}
This work was supported by the Alfred P. Sloan Foundation 
(Grant No.~G-2024-22720) and the National Science Foundation 
(Award No.~2448094).

\subsubsection{\discintname}
The authors have no competing interests to declare that are 
relevant to the content of this article.

\end{credits}

\end{document}